\documentclass[11pt]{article}

\usepackage[final]{acl}

\usepackage{times}
\usepackage{latexsym}

\usepackage[T1]{fontenc}

\usepackage[utf8]{inputenc}

\usepackage{microtype}

\usepackage{inconsolata}

\usepackage{graphicx}

\usepackage{subcaption}
\usepackage{float}
\usepackage{booktabs}
\usepackage{xspace}
\newcommand{\swarm}{\textsc{Swarm}\xspace}

\title{\swarm: A Multilingual Human-Annotated Dataset for Russian Propaganda Detection in Search Engine Results}

\author{
Manuel Tonneau\textsuperscript{1,2} \quad
Abhinav Dubey\textsuperscript{1} \quad
Farhan Shaikh\textsuperscript{1} \quad
Ilaria Vitulano\textsuperscript{1} \\
\bfseries
Martha Stolze\textsuperscript{1} \quad
Hale Dedeoglu\textsuperscript{1} \quad
Clara Riechert\textsuperscript{1} \quad
Ella Kuka\textsuperscript{1} \\
\bfseries
Maryna Sydorova\textsuperscript{3} \quad
Mykola Makhortykh\textsuperscript{3} \quad
Elizaveta Kuznetsova\textsuperscript{1} \\
\\
\textsuperscript{1}Weizenbaum Institute \quad
\textsuperscript{2}University of Oxford \quad
\textsuperscript{3}University of Bern
}

\begin{document}
\maketitle

\begin{abstract}
Russian state propaganda spreads across many languages and online spaces. Yet, most computational work examines only one such space, usually social media, in one or two languages, and analyses sources rather than content. We introduce \textbf{\swarm} (\textbf{S}earch-\textbf{W}eb documents \textbf{A}nnotated for \textbf{R}ussian propaganda, \textbf{M}ultilingual), a dataset of 2{,}183 search engine results across nine languages and diverse web domains (e.g., news, blogs, government sites), each annotated by trained coders for whether it supports a recurring Russian propaganda narrative. We benchmark a source-based blocklist, supervised classifiers, and zero-shot LLMs against these labels. The blocklist misses most propaganda-supporting documents, because such content is not confined to flagged ``propaganda'' outlets but also appears on mainstream ones. Content-level analysis helps, though how much depends on the model: the strongest LLM reaches a positive-class F1 of $0.73$, whereas the supervised classifiers reach only about $0.5$, with the smaller LLMs over-predicting support, mistaking topical relevance for endorsement. Detecting search-borne propaganda thus requires per-language, content-level evaluation, which we hope \swarm and our evaluation code enable.

\end{abstract}

\noindent\textcolor{red}{\textbf{Content warning:} This paper quotes propaganda narratives, including dehumanising elements.}

\begin{figure}[t]
\centering
\includegraphics[width=\columnwidth]{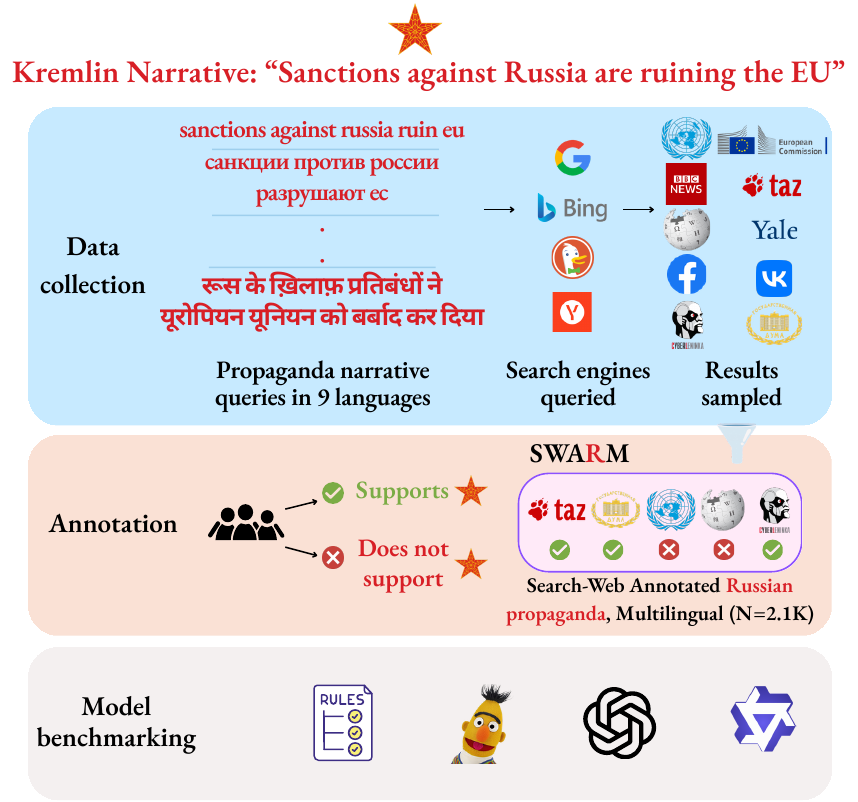}
\caption{Overview of the \textbf{\swarm} pipeline: queries derived from pro-Kremlin narratives (nine languages, four search engines) retrieve documents that annotators label for support of the target narrative, forming the benchmark for narrative-support detection models.}
\label{fig:overview}
\end{figure}

\section{Introduction}
\label{sec:intro}

Propaganda, the systematic manipulation of public opinion to generate support serving the interests of those who deploy it \citep{zollmann_2019}, has long been a tool of the Russian state, intensifying after the 2022 invasion of Ukraine \citep{geissler_russian_2023, matter_temporally_2023}. This manipulation works as intended, with sustained exposure to pro-Kremlin narratives increasing public support for the invasion \citep{krishnarajan_pre-war_2023}.



To counter such manipulation efforts at scale, a substantial body of NLP work targets propaganda detection \citep{da_san_martino_etal_2020}. However, prior work in this area suffers from four major limitations.
First, the \textbf{domain} past work focused on is narrow: existing models focus on a few content types, mainly news and social media, and it is unclear whether they generalise to other domains. Second, \textbf{language coverage} is limited, usually represented by English or Russian alone, whereas Russian propaganda may circulate in many languages. Third, established \textbf{labels} are often unreliable: the few available large-scale datasets rely on weak, source-based labels that assign a document the reputation of its outlet rather than judging its content \citep{barron-cedeno_proppy_2019}. Finally, established \textbf{label schemes} capture which narrative or persuasion technique a document features, not whether it endorses that narrative.

To fill these gaps, we introduce \textbf{\swarm} (\textbf{S}earch-\textbf{W}eb documents \textbf{A}nnotated for \textbf{R}ussian propaganda, \textbf{M}ultilingual), a multilingual dataset for propaganda narrative detection built from an agent-based audit of search engines (\S\ref{sec:data}, Figure~\ref{fig:overview}). \swarm comprises 2{,}183 documents returned by four major search engines for queries on twenty pro-Kremlin narratives, across nine languages (Arabic, English, German, Hindi, Polish, Portuguese, Russian, Spanish, Ukrainian). Trained coders annotate each document for whether it supports a given narrative (e.g., Western sanctions against Russia are ruining Europe's economy), yielding narrative-specific labels rather than broad propaganda-presence judgments. We then benchmark two approaches against these labels (\S\ref{sec:experiments}, \S\ref{sec:results}): a source-based baseline that labels a document by its outlet's reputation, and content-based models that judge the text itself, namely supervised classifiers and zero-shot LLMs. The source-based baseline misses most documents that support a narrative, because such support is not confined to flagged propaganda outlets but also appears on mainstream sources no blocklist includes. Content-based models fare better: zero-shot LLMs outperform the supervised baselines on the minority support class, though the smaller ones systematically over-predict it, conflating topical relevance with endorsement.

In summary, our contributions are:
\begin{itemize}
    \item We release \textbf{\swarm}, the first multilingual, human-annotated dataset for detecting support for Russian propaganda narratives in search engine results, spanning 2{,}183 documents across nine languages (\S\ref{sec:data}), under gated research-use-only access.\footnote{\url{https://huggingface.co/datasets/manueltonneau/SWARM}}
    \item We propose a framework of twenty recurring pro-Kremlin narratives organised around five central actor groups (the United States, ``the West'', the European Union, Russia, and Ukraine), providing a structured basis for narrative-specific support detection in web content (\S\ref{sec:data}).
    \item We benchmark detection approaches against human labels, from a source-based blocklist and supervised classifiers to open and closed zero-shot LLMs (\S\ref{sec:experiments}, \S\ref{sec:results}).
\end{itemize}

\section{Related Work}
\label{sec:related}

\begin{table*}[t]
\centering
\small
\setlength{\tabcolsep}{4pt}
\resizebox{\textwidth}{!}{%
\begin{tabular}{llclllr}
\toprule
Dataset & Domain & \#Lang & Unit & Target label & Annotation & Size \\
\midrule
PTC \citep{da_san_martino_2019} & News & 1 & Fragment & Persuasion technique & Human & 350 \\
SemEval-2023 T3 \citep{piskorski_2023} & News & 9 & Document & Framing, technique & Human & 2{,}049 \\
SemEval-2025 T10 \citep{piskorski_semeval2025_2025} & News & 5 & Document & Narrative typing & Human & 2{,}419 \\
HALT-PROP \citep{rizgeliene_haltprop_2025} & News & 1 & Document & Narrative, technique & Human & 2{,}870 \\
Proppy \citep{barron-cedeno_proppy_2019} & News & 1 & Document & Propaganda (any) & Automatic & 51{,}294 \\
\citet{solopova_propaganda_2023} & News & 5 & Document & Propaganda (any) & Automatic & 18{,}229 \\
HQP \citep{maarouf_hqp_2024} & Social media & 1 & Short text & Propaganda (any) & Human & 30{,}000 \\
\midrule
\textbf{\swarm (ours)} & \textbf{Search results} & \textbf{9} & \textbf{Document} & \textbf{Narrative support} & \textbf{Human} & \textbf{2{,}183} \\
\bottomrule
\end{tabular}%
}
\caption{Prior propaganda datasets and \swarm compared across domain, number of languages, annotation unit, target label, annotation source, and size. Further technique-level datasets are discussed in the text.}
\label{tab:datasets}
\end{table*}

\paragraph{Veracity, framing and narrative} Most computational work on online manipulation targets \emph{veracity}, judging whether a claim is supported, refuted, or unverifiable \citep{guo_2022}. This fits propaganda poorly, which is defined by intent to influence through \emph{framing} and \emph{narrative}, not truth \citep{zollmann_2019, da_san_martino_etal_2020}. Framing makes some aspects of reality salient to promote a problem definition or remedy \citep{entman_1993, otmakhova_2024}, such as spotlighting the economic pain of Western sanctions, whereas narrative orders actors and events to assign causality and blame \citep{bruner_1991, abbott_2021}, such as the story that these sanctions backfired and now punish ordinary Europeans. Because such content can be true, unfalsifiable, or value-laden \citep{wardle_derakhshan_2017}, factual accuracy reveals little about whether a document endorses it. Stance detection is closer \citep{hardalov_2022}, but it targets a text's position toward a single bounded proposition, whereas endorsing a propaganda narrative need not involve any explicit position on one claim. None of these tasks labels whether a document endorses a specific recurring narrative, the target we adopt, which cannot be reduced to topic or sentiment \citep{de_vreese_2005, matthes_kohring_2008}.

\paragraph{Propaganda datasets} Existing datasets use three labelling schemes, none recording narrative support. Most tag the \emph{persuasion technique} of a span \citep{da_san_martino_2019, gupta-etal-2019-neural}. Others label documents by their outlet's propaganda status \citep{rashkin_truth_2017, barron-cedeno_proppy_2019, solopova_propaganda_2023}, which transfers poorly across domains \citep{wang_crossdomain_2020, maarouf_hqp_2024}. A newer line assigns narratives from a fixed taxonomy \citep{piskorski_semeval2025_2025, rizgeliene_haltprop_2025}, which is narrative typing, not support. They also draw mostly on social media \citep{geissler_russian_2023, vanetik_propaganda_2023}, rarely beyond English and Russian \citep{alam_wanlp_2022, chernyavskiy_zenpropaganda_2024}. Even the closest resources, EUvsDisinfo \citep{leite_euvsdisinfo_2024} and RU22Fact \citep{zeng_ru22fact_2024}, label disinformation or veracity, not narrative support. \swarm is the first to provide human, document-level narrative-support labels across many languages and web domains (Table~\ref{tab:datasets}).

\paragraph{Search engines} Search engines are a primary gateway to online information and differ from social media as a channel: the user arrives with an explicit information need, and the ranking carries platform rather than social endorsement. Such engines are also an actively manipulated target: ``data voids'', meaning queries with little authoritative content, can be filled with low-quality material \citep{golebiewski_boyd_2019}, and pro-Kremlin operations have exploited such voids to push their own sources up the rankings \citep{williams_search_2023}. Algorithmic audits confirm the effect, with search engines surfacing substantial misinformation on Russia--Ukraine topics, especially in Russian \citep{kuznetsova_algorithmically_2026}, and this content now contaminating the retrieval and training corpora of LLMs through large-scale ``grooming'' campaigns \citep{dfrlab_pravda_2025}. Despite this reach and exposure to manipulation, propaganda detection work has largely overlooked search engines. Indeed, audits characterise what they \emph{surface} but stop short of \emph{detecting} narrative support in the results. To our knowledge, \swarm is the first to annotate propaganda-narrative support in multilingual search-engine results, judged on content rather than source.

\section{The \swarm{} Dataset}
\label{sec:data}

\swarm is a multilingual dataset for propaganda support detection: search results, each labelled by hand for whether the linked \emph{document} (the web page or PDF the result links to) supports a recurring Russian propaganda narrative. Throughout, \emph{narrative} denotes a recurring element of a pro-Kremlin discourse. To make each narrative checkable, we phrase it as a single yes/no \emph{question}, and consider that a document \emph{supports} the narrative when the answer to that question is yes. This section describes how the dataset is built: query design, audit, sampling, scraping, and annotation.

\subsection{Query corpus}
We define twenty Russian propaganda narratives, organised around five actor groups: the United States, ``the West'', Europe/the EU, Russia, and Ukraine. The narratives are drawn from recurring themes in Russian state media coverage spanning politics, economics, war, pro-Russian/anti-Western values, and anti-gender/LGBTQ values \citep{matter_temporally_2023}. Each narrative is transformed into two search queries, one propaganda-framed and one neutral (for example, ``Western sanctions are ruining Europe's economy'' versus ``effects of Western sanctions on Europe's economy''), and is paired with a single yes/no \emph{question} operationalising the narrative (for example, ``Do sanctions against Russia primarily harm the economy and population of the EU?''). The twenty questions are listed in Appendix~\ref{sec:appendix_claims}. Queries are translated into nine languages by native speakers (Arabic, English, German, Hindi, Polish, Portuguese, Russian, Spanish, and Ukrainian), yielding 40 queries per language and 360 queries in total.

\subsection{Algorithmic audit}
Data are collected through an agent-based algorithmic audit \citep{ulloa_scaling_2024}, in which programmed scripts imitate human browsing behaviour. Using Google Compute Engine, we deploy them across six regions (US/Nevada, India/Mumbai, Qatar/Doha, Brazil/S\~ao Paulo, Poland/Warsaw, and Germany/Frankfurt), each running two browsers (Firefox and Chrome). Every query is issued in all six regions, so each language is observed across all locations. Agents issue every query to Google, Bing, Yandex, and DuckDuckGo through the text-search interface on 29--30 January 2024. The audit yields over 1.3 million result listings, comprising 30{,}265 unique URLs from 7{,}191 unique domains.

\subsection{Sampling}
We draw a stratified sample of the collected results: for each language, 15 results per narrative, drawn from both the propaganda-framed and the neutral query for that narrative, yielding \textbf{300 documents per language} (2{,}700 across the nine languages). Because 15 is odd, the two framings cannot be split evenly within a narrative. Across the full sample, 53\% of results come from propaganda-framed queries and 47\% from neutral ones, with the propaganda-framed share per language ranging from 46\% (Hindi) to 59\% (Ukrainian). The sample spans the search engines and regions of the audit, making it a designed sample of real search result environments rather than a convenience collection.

\subsection{Scraping}
Each sampled page is crawled and its main text extracted, using \texttt{trafilatura} \citep{barbaresi-2021-trafilatura} for HTML pages and \texttt{pdfplumber}\footnote{\url{https://github.com/jsvine/pdfplumber}} for PDF documents. A URL enters the dataset only if a clean body of text can be extracted and then assessed by annotators. Coders and models judge the full document text, not the snippet shown in the search engine result list. As expected from a large-scale web collection, a share of URLs yields no usable text, due to paywalls, bot protection, client-side rendering, dead links, and blocked or unresolvable domains, and a recovery pass (live re-fetch with a Wayback Machine\footnote{\url{https://web.archive.org/}} fallback) restores part of that shortfall (Appendix~\ref{sec:appendix_retrieval}). After annotation, 2{,}183 of the 2{,}700 sampled rows are ultimately retained, with retention varying across languages (Table~\ref{tab:perlang}).

The 517 sampled rows not retained correspond to 503 distinct URLs, since some URLs were sampled twice. Extraction failure accounts for 293 of them, counting pages that return a response but no document, such as paywall and login interstitials, a further 204 were annotated but their extracted text could not be assessed, and 6 ended in a label tie (Appendix~\ref{sec:appendix_retrieval}). Extraction does not succeed uniformly across sources: Russian state-linked domains were over-represented among the initial failures, but the recovery pass removed most of that skew, leaving the non-retained residue only mildly enriched in such sources (18\% against 15\%, Appendix~\ref{sec:appendix_retrieval}). The retained set may therefore still slightly under-count the positive class, so the support rates we report are, if anything, a lower bound.

\begin{table}[t]
\centering
\small
\begin{tabular}{lccc}
\toprule
Lang & N & Pos.\ rate & $\alpha$ \\
\midrule
EN & 226 & 20\% & 0.41 \\
RU & 248 & 36\% & 0.62 \\
ES & 240 & 14\% & 0.47 \\
PT & 246 & 17\% & 0.59 \\
DE & 257 & 21\% & 0.47 \\
HI & 218 & 8\% & 0.56 \\
PL & 244 & 10\% & 0.44 \\
AR & 258 & 18\% & 0.53 \\
UK & 246 & 15\% & 0.60 \\
\midrule
\textbf{Total} & \textbf{2{,}183} & & \textbf{0.54} \\
\bottomrule
\end{tabular}
\caption{Per-language composition of the labelled set: number of labelled documents (\emph{N}), share of the positive (propaganda-supporting) class (\emph{Pos.\ rate}), and nominal inter-annotator agreement (Krippendorff's $\alpha$).}
\label{tab:perlang}
\end{table}

\subsection{Annotation}
Trained annotators label the sampled documents in Label Studio\footnote{\url{https://labelstud.io/}} following the codebook in Appendix~\ref{sec:appendix_guidelines}. They judge the English translation of the extracted document text, machine-translated with \texttt{gpt-5-nano} through the Batch API, which is also the input given to the English-input models, so coders and models see the same evidence. Working in one language lets a coder handle any of the nine, and it removes translation as a difference between the human and model conditions. We check the translations both directly, through per-language adequacy ratings by native or fluent speakers, and indirectly, by rerunning every model on the source-language text (Appendix~\ref{sec:appendix_translation}). Each document receives a binary label: 1 if it supports the narrative its query was drawn from and 0 otherwise, where the negative class covers documents that debunk or deny the narrative as well as ones that are neutral, two-sided, or unrelated. Documents that cannot be judged, because the link does not open or the extracted text is unusable, are marked ``cannot assess'' and excluded from the released set.

\paragraph{Coding protocol} Each document is labelled independently by two annotators, with a third adjudicating where the two disagree, so the gold label is the majority vote over usable votes and ties are dropped. A document enters the dataset only if at least one coder could assess that text. Polish items were re-coded after an initial pass was found to over-flag the positive class, and the re-coded labels replace it throughout. The released set comprises 2{,}183 labelled documents used in all experiments (Table~\ref{tab:perlang}).

\paragraph{Inter-annotator agreement} Because the number of coders per item varies (two, or three after adjudication), we measure reliability with Krippendorff's $\alpha$. Pooled across languages, agreement reaches $\alpha = 0.54$, ranging from $0.41$ (English) to $0.62$ (Russian), which is moderate (Table~\ref{tab:perlang}). For English and Russian it is computed on the subset of items with per-annotator labels (52 and 70 respectively). The third coder adjudicates a disagreement rather than coding independently, so $\alpha$ here is not a pure independent-coder estimate.

\paragraph{What coders disagree about} Adjudication was needed on 16\% of double-coded items, and the disagreements are far from evenly spread across narratives. They concentrate on evaluative narratives whose supporting evidence is a matter of interpretation, such as \emph{``is the Russian economy stable and able to prosper despite Western sanctions?''} (25\%) and \emph{``do Western media censor important information?''} (24\%), and are rare on narratives that make a concrete, checkable assertion, such as \emph{``is the Ukrainian army weak because homosexuals are part of it?''} (4\%) and \emph{``does the Ukrainian army have Nazi ideology?''} (5\%). When coders do disagree, the adjudicator sides with \emph{not support} in 62\% of cases, so the characteristic human error is to read endorsement into a document that raises a narrative without endorsing it. This is the same failure the models show (\S\ref{sec:results}), which suggests the boundary between raising and endorsing a narrative is intrinsically hard rather than an artefact of our annotation instructions.

\begin{table*}[t]
\centering
\small
\begin{tabular}{p{3.5cm} p{10.5cm} c}
\toprule
Propaganda question & Supporting excerpt (English translation) & Src \\
\midrule
Do the USA strive for global dominance? & ``The United States uses the dollar as a lever of economic and political pressure. When countries begin to abandon the dollar [...] America's world hegemony will waver.'' & UK \\
\addlinespace
Do Western media censor important information? & ``Censorship in Western media is growing, against everything that does not conform to the narrative of the elites. By blocking RT, the EU aims to silence a critical, non-Western-aligned information source.'' & DE \\
\addlinespace
Do sanctions against Russia primarily harm the economy and population of the EU? & ``Sanctions on Russia also harm the EU and increase the East's power [...] partnerships with China and other countries [...] strengthen the Kremlin's position.'' & PT \\
\addlinespace
Does Russia protect children against gay propaganda? & ``The whole media establishment is up in arms about Russia's new law that prohibits the promotion of homosexuality [...] among minors [...] but this appears to be more hyperbole than anything else.'' & EN \\
\addlinespace
Does the Ukrainian army have Nazi ideology? & ``The New York Times palms off the deep historical and present-day links of Ukrainian nationalism to Nazism and genocide as merely `thorny issues' [...] a public relations problem for media propagandists.'' & EN \\
\bottomrule
\end{tabular}
\caption{Examples of search-retrieved content \emph{supporting} a Russian propaganda narrative, one per actor group, in the English translation given to annotators and models. \emph{Src} is the source language.}
\label{tab:examples}
\end{table*}

\subsection{Dataset description}
Each \swarm record pairs a retrieved document (its URL, title, source language, and extracted English text) with the narrative it is retrieved for, the issuing query and its framing, and the binary support label.

\paragraph{Examples} Table~\ref{tab:examples} shows representative search-retrieved documents that support a propaganda narrative, one per actor group, with a short excerpt from each.

\paragraph{Descriptive statistics} Of the 2{,}183 labelled documents, 388 (18\%) are positive, that is, they support a propaganda narrative. The positive class is the minority in every language, from 8\% (Hindi) to 36\% (Russian), with the per-language counts and positive rates in Table~\ref{tab:perlang}.

\paragraph{Source composition} To characterise the domains the dataset covers, we assign each document's source domain to one of nine categories: news outlet, blog, research institute, government or state site, international organisation, social media platform, fact-checker, Wikipedia, and other. Curated lists and a news-reliability list settle most domains, and \texttt{gpt-5-nano} labels the remainder (92\% document-weighted accuracy, Appendix~\ref{sec:appendix_domain_validation}). The corpus is overwhelmingly \emph{news and web media}: news outlets account for 60\% of documents, followed by blogs and research institutes (9\% each), \emph{other} sites (6\%), international organisations and Wikipedia (5\% each), government/state sites (4\%), social media (2\%), and fact-checkers (under 1\%, Table~\ref{tab:source_types}). The two shares we draw conclusions from are the most robust: news outlets are categorised at 92\% accuracy, and the 2\% social media share is if anything an over-estimate.

Two points stand out. First, only 2\% of search results link to social media: unlike prior corpora built from social media posts, \swarm captures narrative support as it surfaces in the open web. Second, propaganda support varies by source type, highest on social media (33\%) and blogs (25\%), around 20\% on news and government sites, and lowest on international organisation and Wikipedia pages (3\%, Table~\ref{tab:source_types}). The blog figure needs more caution, since \emph{blog} is among the least reliably assigned categories.

\section{Experiments}
\label{sec:experiments}

We now set up our benchmark: the models we compare and the evaluation protocol. The task is binary: each model reads a document together with a yes/no \emph{question} and predicts whether the document supports the narrative (1) or not (0), following the annotation codebook (Appendix~\ref{sec:appendix_guidelines}). All models are run on the full labelled set of 2{,}183 documents across the nine languages, so every method is scored on exactly the same items, with the supervised baselines evaluated out of fold rather than on a held-out split.

\subsection{Models}

We benchmark models from three families: a source-based domain baseline, three supervised baselines (logistic regression on frozen embeddings, XLM-R, and SetFit), and four zero-shot LLMs (the open Qwen2.5-7B and Qwen2.5-72B and the closed GPT-5-nano and GPT-5.4).

\paragraph{Source-based domain baseline} Drawing on the source-based labelling shortcut criticised in \S\ref{sec:related}, we predict support whenever a document's registered domain appears on a curated list of Russian propaganda websites. The list merges four public propaganda-domain resources, matched at the registered-domain level: a scraped Wikipedia list of Russian disinformation sites, EUvsDisinfo \citep{leite_euvsdisinfo_2024}, Proppy/MBFC \citep{barron-cedeno_proppy_2019}, and the corpus of \citet{rashkin_truth_2017}.

\paragraph{Supervised} We train three supervised models on our labels with 5-fold cross-validation stratified by language and label, scoring every item out of fold. All three encode only a document's first 512 tokens:
\begin{itemize}
\item \textbf{Embeddings + logistic regression}: we fit an L2-regularised logistic regression on frozen \texttt{multilingual-e5-large} \citep{wang_multilingual_e5_2024} sentence embeddings.
\item \textbf{XLM-R}: we fine-tune an XLM-RoBERTa-large encoder \citep{conneau_unsupervised_2020}, the standard multilingual baseline.
\item \textbf{SetFit}: we apply the sample-efficient few-shot method of \citet{tunstall_setfit_2022}, contrastively fine-tuning \texttt{multilingual-e5-large} on same- vs.\ different-label sentence pairs and then fitting a lightweight classifier head on the resulting embeddings. It needs far fewer labels than full fine-tuning, which suits our 388 positives spread over nine languages.
\end{itemize}

\paragraph{Zero-shot LLMs} We evaluate four LLMs zero-shot, two open and two closed. The \textbf{open} models are Qwen2.5-7B-Instruct and Qwen2.5-72B-Instruct \citep{qwen2.5}, chosen for the family's strong coverage of our nine languages and run at two sizes to span a small-versus-large axis. The \textbf{closed} models are GPT-5-nano, a small low-cost model of the kind likely deployed at audit scale, and GPT-5.4, a frontier state-of-the-art reference. Both run at high reasoning effort. GPT-5-nano also produces the English translations (\S\ref{sec:data}) and the residual domain categorisation (Appendix~\ref{sec:appendix_domain_validation}), so it is both a tool and a system under test. Running every model on untranslated text (\S\ref{sec:experiments}) provides a check free of that dependency. Every model receives the same fixed prompt, holding the yes/no question and the document truncated to 12{,}000 characters, and returns a single binary judgment. The prompt, its provenance, and the inference settings are in Appendices~\ref{sec:appendix_prompt} and \ref{sec:appendix_hyperparams}, with a prompt ablation in Appendix~\ref{sec:appendix_ablation}.

\begin{table*}[t]
\centering
\small
\begin{tabular}{l c @{\hspace{1.6em}} ccc @{\hspace{1.6em}} cccc @{\hspace{1.6em}} c}
\toprule
 & Rule-based & \multicolumn{3}{c}{Supervised} & \multicolumn{4}{c}{Zero-shot} & \\
\cmidrule(lr){2-2}\cmidrule(lr){3-5}\cmidrule(lr){6-9}
Lang & Domain & XLM-R & e5-LR & SetFit & Qwen-7B & Qwen-72B & GPT-5-nano & GPT-5.4 & Best \\
\midrule
AR & 0.31 & 0.48 & 0.45 & 0.37 & 0.60 & 0.62 & \textbf{0.67} & 0.65 & 0.67 \\
DE & 0.07 & 0.48 & 0.43 & 0.37 & 0.59 & 0.60 & \textbf{0.76} & 0.68 & 0.76 \\
EN & 0.38 & 0.46 & 0.45 & 0.44 & 0.62 & 0.69 & 0.72 & \textbf{0.74} & 0.74 \\
ES & 0.46 & 0.49 & 0.38 & 0.50 & 0.53 & 0.54 & 0.59 & \textbf{0.65} & 0.65 \\
HI & 0.04 & 0.34 & 0.36 & 0.52 & \textbf{0.81} & 0.76 & 0.73 & 0.73 & 0.81 \\
PL & 0.22 & 0.44 & 0.36 & 0.49 & 0.51 & 0.55 & 0.61 & \textbf{0.67} & 0.67 \\
PT & 0.23 & 0.55 & 0.54 & 0.64 & 0.58 & \textbf{0.69} & 0.58 & 0.60 & 0.69 \\
RU & 0.64 & 0.67 & 0.64 & 0.68 & 0.78 & 0.79 & 0.79 & \textbf{0.82} & 0.82 \\
UK & 0.50 & 0.43 & 0.41 & 0.44 & 0.54 & 0.59 & 0.71 & \textbf{0.75} & 0.75 \\
\midrule
\textbf{All} & 0.39 & 0.51 & 0.47 & 0.51 & 0.62 & 0.66 & 0.70 & \textbf{0.71} & 0.73 \\
\bottomrule
\end{tabular}
\caption{Positive-class F1 per language and overall, English input, on the 2{,}129-document labelled set, best per row in bold. \emph{Domain} is the non-learned source-based blocklist, XLM-R, e5-LR, and SetFit are the supervised baselines, and the remaining columns are the zero-shot LLMs. \emph{Best} is the highest F1 any method reaches for that language. Its \emph{All} entry is the \emph{oracle}, the mean of these per-language ceilings.}
\label{tab:per_lang_posf1}
\end{table*}

\subsection{Evaluation}

\paragraph{Metrics} We report positive-class F1 given the rare support class, with bootstrap 95\% confidence intervals over items (2{,}000 resamples). Precision, recall, and balanced accuracy are in Appendix~\ref{sec:appendix_results}. Because the support class is small and unevenly distributed across languages (Table~\ref{tab:perlang}), we report F1 both per language and overall, since aggregates can mask language-dependent failures.

\paragraph{Input language} We evaluate on two input variants: the English machine translation of each document, which is also the text shown to annotators, and the native source-language text. All supervised and zero-shot models are run on both variants. The source-based baseline does not read document text. Translation adequacy is assessed in Appendix~\ref{sec:appendix_translation}.

\section{Results}
\label{sec:results}

\paragraph{Source-based labels are not enough} We first ask whether support can be read off a document's \emph{source} alone, using the domain blocklist of \S\ref{sec:experiments}. Across all 2{,}183 documents, the blocklist reaches a positive-class F1 of only $0.39$ (Table~\ref{tab:full_results} in Appendix~\ref{sec:appendix_results}). It flags 15\% of documents and recovers just 35\% of supporting documents at 43\% precision. Its coverage is also concentrated on news: it recovers 41\% of support on news outlet pages but only 22\% on blogs and 3\% on research institute pages (Table~\ref{tab:source_types}). The misses are not merely a coverage gap: of the 251 supporting documents the blocklist overlooks, 69\% appear on mainstream news, government, international organisation, or fact-checker domains (Table~\ref{tab:domain_recall} in Appendix~\ref{sec:appendix_results}), where propaganda narratives are amplified without the outlet itself being ``propaganda''. One such miss is a commentary in the German daily \emph{taz} arguing that sanctions fail to hurt Russia while acting as a ``boomerang'' on Europe, a recurring pro-Kremlin narrative on an outlet no blocklist would flag. As the Ethics Statement notes, this labels a single commentary, not the outlet's editorial line. Detecting narrative support thus requires judging documents individually, the task we benchmark next.

\paragraph{LLMs beat supervised} The three supervised methods reach overall F1 scores of only 0.47--0.51 (e5-LR 0.47, XLM-R 0.51, SetFit 0.51), against 0.62--0.71 for the zero-shot LLMs. SetFit shows why accuracy is misleading here: its raw accuracy is high (0.85, matching Qwen2.5-72B and behind only GPT-5.4 at 0.90 and GPT-5-nano at 0.88, Table~\ref{tab:full_results}) but with only 388 positives, it favours the majority class, recovering just 0.44 of support cases against 0.76--0.79 for the open and small closed LLMs. The supervised methods are competitive only in a few languages (SetFit reaches 0.68 on Russian and 0.64 on Portuguese), yet in every language the best zero-shot LLM still outscores the best supervised method, by margins from 0.05 (Portuguese) to 0.31 (Ukrainian).

\paragraph{No clear LLM winner} GPT-5.4 reaches the best overall F1 (0.71), ahead of the small GPT-5-nano (0.70) and the open Qwen2.5-72B (0.66), but the margin is small and no model is reliably best across languages (Table~\ref{tab:per_lang_posf1}): GPT-5-nano wins Arabic and German, Qwen2.5-7B wins Hindi, and Qwen2.5-72B wins Portuguese, so a strong open model is competitive with small closed ones. Per-language winners should nonetheless be read with caution, since several languages carry few positives, as few as 17 in Hindi (Table~\ref{tab:perlang}). Bootstrap resampling still finds these winners mostly stable, the exception being Arabic, whose ranking is unreliable (Appendix~\ref{sec:appendix_results}). This variation is also hard to exploit: an oracle selecting, for each language, the best of the seven models and two input variants averages 0.74 F1, only 0.01 above running GPT-5.4 on native input throughout (0.73), so tailoring the choice to each language yields a negligible gain.

\paragraph{The translation effect} Input language matters little: moving each model from the English translation to the native source-language text shifts positive-class F1 by at most 0.02 in either direction. The open zero-shot LLMs lose slightly on native input (Qwen2.5-7B 0.62 to 0.60, Qwen2.5-72B 0.66 to 0.64), the closed models move in opposite directions (GPT-5-nano 0.70 to 0.69, GPT-5.4 0.71 to 0.73), and the supervised baselines mostly gain (SetFit 0.51 to 0.53, e5-LR 0.47 to 0.48, XLM-R 0.51 to 0.49). Translation therefore helps the open LLMs modestly but neither the closed LLMs nor the encoders, which handle the source text at least as well (Table~\ref{tab:full_results} in Appendix~\ref{sec:appendix_results}).

\paragraph{Error analysis} The smaller LLMs share one failure mode: \emph{over-prediction of narrative support}. Taking GPT-5-nano as representative, recall is high (0.76) but precision lower (0.65), with more false positives than false negatives. False positives concentrate on narratives whose topic a document can discuss factually without endorsing the propaganda framing, most frequently \emph{``does Russia protect children against gay propaganda?''}, \emph{``is the gender debate an agenda of the European elites?''}, and \emph{``do Western media censor important information?''}. The models tend to treat on-topic documents as supportive. False negatives are fewer and concentrate on economic narratives where support is implicit, such as \emph{``is the Russian economy stable and able to prosper despite Western sanctions?''}. SetFit fails in the opposite direction, under-predicting support (precision 0.61, recall 0.44), and the two encoder baselines (XLM-R, e5-LR) instead over-predict like the small LLMs. GPT-5.4 is the most precise (precision 0.73, recall 0.69), which is why it leads on F1 despite a lower recall than the smaller models. Overall, the boundary between neutral coverage and endorsement on a shared topic is where current systems struggle.

\section{Discussion}
\label{sec:discussion}

\paragraph{Source is not a proxy for content} Our results confirm, in a multilingual search-result setting, the critique that source-based labels diverge from human content judgments \citep{barron-cedeno_proppy_2019, maarouf_hqp_2024}, and extend it: narrative support is amplified through ordinary media that no blocklist would contain \citep{williams_search_2023, kuznetsova_algorithmically_2026}, so source is not a usable proxy for content.

\paragraph{Zero-shot LLMs beat supervised baselines} That zero-shot LLMs beat every supervised baseline on the support class is counterintuitive: with in-domain labels, fine-tuned encoders usually match or exceed them \citep{sprenkamp_2023, hasanain_2024, tonneau-etal-2025-hateday}. Two features of our setting drive the reversal. Support is rare, so with only 388 positives the encoders see too few examples to learn a robust boundary while the LLMs fall back on pre-training knowledge, and the encoders read only the first 512 tokens, missing support expressed further down. Capping the LLMs at the same 512-token budget (Appendix~\ref{sec:appendix_context}) costs Qwen2.5-7B 0.03 F1 and Qwen2.5-72B 0.01, yet both stay clearly ahead of the best supervised baseline, so label scarcity, not context length, is the main factor. More annotated data would give supervised methods a fairer test.

\paragraph{Topical relevance is not endorsement} The systems' dominant error is to flag any on-topic document as supportive, failing at the boundary between raising a narrative and endorsing it. This matches the view that propaganda is defined by framing rather than veracity \citep{da_san_martino_etal_2020}. It also matches the gap left by narrative-typing datasets, which record which narrative a document raises but not whether it endorses it \citep{piskorski_semeval2025_2025, rizgeliene_haltprop_2025}, the distinction our support label targets. Prompts or fine-tuning that make this boundary explicit, for instance by asking for the passage that endorses the narrative, are therefore the most direct route to reducing the over-prediction we observe.

\paragraph{Language disparities} No system is reliably best across languages. On the prevalence-robust balanced accuracy, English is not the best-detected language: its per-language ceiling (0.83) trails Polish (0.87), Russian (0.86), German and Ukrainian (0.85), and Portuguese (0.84, Table~\ref{tab:per_lang_balacc}), as well as Hindi, whose 0.98 rests on only 17 positives and should be read with caution. This runs against the finding that generative LLMs are strongest in English \citep{ahuja_mega_2023}, while the strength on Russian and Ukrainian fits their proximity to the propaganda's origin and the language-dependent LLM behaviour reported on such content \citep{makhortykh_stochastic_2024, urman_silence_2024}. The disparities are not an artefact of translation, since native-input scores move by at most 0.02 (\S\ref{sec:results}). Per-language, narrative-level reporting is therefore needed rather than a single aggregate score.

\paragraph{Implications for measurement and future work} A frontier model reaches 0.73 positive-class F1 on average, respectable for a hard multilingual task yet short of reliable, so downstream estimates of how much propaganda search engines surface should correct for classifier error rather than treat predicted labels as ground truth \citep{egami_imperfect_2023}. The benchmark also leaves clear room for better detectors. Because we fixed the zero-shot prompt in advance (\S\ref{sec:experiments}), the scores are conservative, and requiring a short rationale or supporting quote before the label, adding self-consistency or probability calibration, or parameter-efficient fine-tuning on the \swarm labels are natural routes to curbing the over-prediction we observe. Two ablations point the same way: removing the prompt's self-consistency instruction moves F1 by under 0.01 and capping the LLMs at 512 tokens costs at most 0.03 (Appendices~\ref{sec:appendix_ablation} and \ref{sec:appendix_context}), so the headroom lies in the task formulation rather than in decoding or input budget. A further direction is to test whether detectors trained on pro-Kremlin narratives transfer to other state actors, which would separate what is specific to this corpus from what is general to propaganda narrative-support detection.

\section{Conclusion}
\label{sec:conclusion}
We introduced \swarm, the first multilingual, human-annotated dataset for detecting support for Russian propaganda narratives in search engine results, and benchmarked source-based, supervised, and zero-shot approaches against its labels. Source-level analysis fails on this task, missing most narrative support, much of it from mainstream outlets. Zero-shot LLMs outperform the supervised baselines on the rare support class but remain far from reliable, the smaller ones over-predicting support by mistaking topical relevance for propaganda endorsement. Improving content-level, per-language detection, through task-tailored prompting, fine-tuning on in-domain labels such as ours, or precision-oriented decision rules, is therefore a central goal for future work. Because search-borne propaganda shapes public opinion and feeds the training corpora of language models, reliable detection is a prerequisite for auditing what search engines surface and filtering what models ingest, and we hope \swarm, released with its guidelines and evaluation code, drives progress toward both.

\section*{Limitations}

\paragraph{Retrieval selection bias} A document enters the benchmark only if its page could be retrieved as clean text. Retrieval failures (paywalls, bot protection, dead or blocked domains) initially skewed toward Russian state outlets, several unresolvable from our collection setup. A targeted recovery pass restored most such pages, and in the released set the non-retained residue is only mildly skewed toward pro-Kremlin sources (18\% of non-retained results fall on our propaganda blocklist against 15\% of the retained set, Appendix~\ref{sec:appendix_retrieval}). Any remaining under-representation of overtly propaganda-supporting sources is therefore small.

\paragraph{Label scarcity and construct} The positive (support) class is rare (18\% overall, as low as 8\% in some languages), so positive-class metrics carry wide confidence intervals and per-language scores for the smallest counts (e.g.\ Hindi) should be read with caution. ``Support for a propaganda narrative'' is also a contestable construct: it is operationalised here through fixed yes/no questions and a binary label, which cannot capture partial endorsement, irony, or framing effects. Chance-corrected agreement is accordingly moderate and uneven across languages.

\paragraph{Data age and duplication} The audit was run on 29--30 January 2024, so the corpus is a snapshot, and both the narratives in circulation and the engines' ranking will have moved since. Separately, 40 documents (1.8\% of the set) fall into 16 near-duplicate groups, 2 of which carry conflicting gold labels. We document rather than filter them, so a small amount of leakage across cross-validation folds is possible for the supervised baselines, though at this scale it cannot account for the gap to the zero-shot models.

\paragraph{Scope of the benchmark} We evaluate three supervised methods and four zero-shot LLMs (two open, two closed), leaving other frontier models and fine-tuning of LLMs to future work. The supervised baselines are further limited by the scarce positives and by 512-token truncation, so they see only the opening of each document, whereas the LLMs read far more of it. Search results are also a snapshot of six specific regions and a two-day window, and LLM detectors carry their own language-dependent biases \citep{urman_silence_2024}, a further reason to report per-language rather than aggregate scores.


\section*{Ethics Statement}

\paragraph{Annotation} Documents were labelled by members of the research team and collaborating volunteers who are native or fluent speakers of the respective languages, not by paid crowdworkers. Participation was voluntary and informed. Because annotators were exposed to propaganda and at times dehumanising content, they could pace their work, skip items, and stop at any time. Labels record whether a document \emph{supports a propaganda narrative}. They are not judgments of the personal views of any author or publisher, and a positive label does not imply that an outlet endorses the narrative as a matter of editorial position.

\paragraph{Data release} The documents originate from public search engine results and may be subject to the source engines' terms of service and to third-party copyright. To respect these constraints, and because a detector of narrative support is dual-use in that the same labels could in principle be used to tune content that \emph{evades} detection, the dataset is released under gated, research-use-only access with a request process and a takedown procedure (see the Data and Code Availability section). Quoted examples in the paper are kept to a minimum and stripped of information that could re-identify individuals. We follow the data statement framework of \citet{bender_friedman_2018}, given in Appendix~\ref{sec:appendix_datastatement}, and community guidance on handling and presenting harmful text \citep{kirk_handling_2022}.


\section*{Author Contributions}

Manuel Tonneau led the project and contributed to data processing, coordination of the annotation process, data analysis, modeling, and writing. The study was designed by Manuel Tonneau, Abhinav Dubey, Farhan Shaikh, Ilaria Vitulano, Martha Stolze, Mykola Makhortykh, and Elizaveta Kuznetsova. Maryna Sydorova collected the data. Abhinav Dubey and Farhan Shaikh supported the modeling. Annotation was carried out by Ilaria Vitulano, Martha Stolze, Hale Dedeoglu, Clara Riechert, Ella Kuka, Abhinav Dubey, and Farhan Shaikh. Mykola Makhortykh and Elizaveta Kuznetsova supervised the project. Ilaria Vitulano, Martha Stolze, Mykola Makhortykh, and Elizaveta Kuznetsova provided critical feedback on the manuscript.

\section*{Data and Code Availability}
\label{sec:availability}

\swarm is available at \url{https://huggingface.co/datasets/manueltonneau/SWARM} under gated, research-use-only access. Requests are reviewed individually, and access requires agreeing to use the data for non-commercial research, not to redistribute it, and to honour takedown requests from rights holders. The release contains all 2{,}183 annotated documents with their labels, the narrative question each was annotated against, and both the English-translated and the source-language text used as model input. 

The evaluation code is available at \url{https://github.com/manueltonneau/swarm}. It covers the source-based blocklist, the supervised baselines, the zero-shot runners for both open and closed models, and the analysis that produces every table reported here, together with a script that turns the released dataset into the exact inputs used for each experiment.

\section*{Acknowledgements}

We thank Tamer Farag, Alessandra Ciccolella and Nina Tsarskaia for their contributions to the annotation of this dataset.

This publication was supported by the Weizenbaum Institute (grant number 16DII141), funded by the Federal Ministry of Research, Technology and Space (BMFTR) and the State of Berlin.

\bibliography{custom}

\appendix

\section{Annotation Codebook}
\label{sec:appendix_guidelines}

\paragraph{Study overview} The data come from a study of information environments on search engines. Using an agent-based audit, we gather search results from Google, Bing, DuckDuckGo, and Yandex in nine languages (English, Spanish, German, Portuguese, Arabic, Polish, Hindi, Russian, and Ukrainian). The aim is to study how search engines handle Russian propaganda by comparing the results returned for neutral queries against those returned for queries framed using common Russian propaganda narratives. Trained coders annotate a subsample by hand to provide the human labels against which automatic detection methods are evaluated. The data sent to annotators are not pre-labelled by any model.

\paragraph{Item structure} Each item provides: the link to a document appearing in the search result, the document title, the query used for the search, the propaganda counterpart of the query (equal to the query when the query is already propaganda), and a yes/no \emph{question} whose ``yes'' answer means the document supports the Russian propaganda narrative.

\paragraph{Task} For each item, the annotator reads the question and the document text and identifies the passage related to the query (a page may contain several unrelated pieces, and only the relevant one is judged). The item also carries the source link, which annotators could open when the extracted text was unclear. Annotators read with the question in mind and look for the answer, spending roughly two minutes per document and not needing to read a long document in full once the answer is clear. Unusable text, and any uncertainty or comments, are recorded in dedicated fields.

\paragraph{Label} Annotators assign a binary label. Items whose text could not be assessed are left blank and excluded.

\begin{itemize}
    \item \textbf{1 -- supports the propaganda narrative.} The document clearly and unequivocally answers ``yes'' to the question. The narrative is presented as fact and is neither debunked nor critically discussed (for example, quoting Russian officials or documents without context or counterargument), or the document provides evidence or analysis supporting it. It may be carried by a non-Russian or otherwise reputable source: judge the content, not the source.
    \item \textbf{0 -- does not support the propaganda narrative.} The document either clearly denies or debunks it (including subtle forms such as irony or scare quotes), or it neither supports nor denies the narrative, presents both sides without a conclusion, or is unrelated to the propaganda narrative.
    \item \textbf{blank -- cannot assess.} The extracted text is unusable or the document is unavailable. Excluded from the evaluation set.
\end{itemize}

\paragraph{Examples} Table~\ref{tab:codebook_examples} lists example annotations under this scheme.

\begin{table}[t]
\centering
\small
\begin{tabular}{p{5.4cm}c}
\toprule
Article title (query) & Label \\
\midrule
Russia among leaders of process of building multipolar world \emph{(russia multipolar world fair)} & 1 \\
\addlinespace
Discrimination against the Russian language in Ukraine \emph{(ukraine russian language discrimination)} & 1 \\
\addlinespace
A `multipolar' world defies the `rules-based' order \emph{(russia multipolar world fair)} & 0 \\
\addlinespace
The West must abandon weakness and commit to Ukraine's victory \emph{(ukraine army gays weak)} & 0 \\
\addlinespace
Homophobic fake news circulates about recruitment for the Ukrainian army \emph{(ukraine army gays weak)} & 0 \\
\addlinespace
Russian invasion of Ukraine \emph{(russia ukraine invasion)} & 0 \\
\bottomrule
\end{tabular}
\caption{Example annotations under the binary scheme.}
\label{tab:codebook_examples}
\end{table}

\section{Translation Quality}
\label{sec:appendix_translation}

Document text was machine-translated to English with \texttt{gpt-5-nano} (Batch API) for both annotation and the English-input models (\S\ref{sec:data}). We check adequacy in two ways. \emph{Directly}, one native or fluent speaker per language rated a stratified sample of about ten documents per source language (78 items across the eight non-English languages). Each item was rated once, so this is a spot check rather than a reliability-controlled measurement. Ratings were given on a three-point scale of \emph{adequate} (all meaning preserved), \emph{minor loss} (some nuance lost), and \emph{inadequate} (meaning materially distorted or wrong), checking whether the main assertions and stance survived. As Table~\ref{tab:translation_quality} shows, \textbf{86\% of translations were rated adequate, 14\% minor loss, and none inadequate}. Ratings are perfect for Spanish, Hindi, Portuguese, and Russian and softest for German (60\%) and Polish (70\%), where the losses were nuance rather than meaning. \emph{Indirectly}, evaluating each model on the source-language text rather than the English translation changes positive-class F1 by at most 0.02 and leaves the two closed models essentially unchanged. Translation gives the open zero-shot LLMs a small, consistent gain and does not help the closed models or the encoder baselines, which handle the source text at least as well. Together, the direct ratings and the small downstream gaps indicate that translation preserves the content needed for narrative-support detection rather than introducing systematic distortion.

\begin{table}[h]
\centering
\small
\begin{tabular}{lrrr}
\toprule
Lang & $n$ & Adequate & Minor loss \\
\midrule
AR & 10 & 8 & 2 \\
DE & 10 & 6 & 4 \\
ES & 10 & 10 & 0 \\
HI & 9 & 9 & 0 \\
PL & 10 & 7 & 3 \\
PT & 10 & 10 & 0 \\
RU & 9 & 9 & 0 \\
UK & 10 & 8 & 2 \\
\midrule
\textbf{All} & \textbf{78} & \textbf{67} & \textbf{11} \\
\bottomrule
\end{tabular}
\caption{Human translation-adequacy ratings on a stratified sample (${\sim}10$ documents per source language). No translation was rated \emph{inadequate}, so that column is omitted.}
\label{tab:translation_quality}
\end{table}

\section{Propaganda Narratives and Questions}
\label{sec:appendix_claims}

The twenty narratives used as annotation targets, each stated as the yes/no question put to annotators and models, grouped by actor group. Every narrative is issued to the search engines as both a propaganda-framed and a neutral-framed query.

\paragraph{United States}
\begin{enumerate}\itemsep0pt
    \item Do the USA strive for global dominance?
    \item Is the US economy collapsing due to the financial policy of the US elites?
\end{enumerate}
\paragraph{The West}
\begin{enumerate}\itemsep0pt \setcounter{enumi}{2}
    \item Do Western media censor important information?
    \item Is the representation of transgender people in public office a sign of moral decline of Western society?
\end{enumerate}
\paragraph{European Union}
\begin{enumerate}\itemsep0pt \setcounter{enumi}{4}
    \item Is the EU a vassal of the US?
    \item Do sanctions against Russia primarily harm the economy and population of the EU?
    \item Does European foreign policy have moral double standards?
    \item Is the gender debate an agenda of the European elites?
    \item Does Europe fuel the war in Ukraine with arms deliveries?
\end{enumerate}
\paragraph{Russia}
\begin{enumerate}\itemsep0pt \setcounter{enumi}{9}
    \item Does Russia advocate for a fair multipolar world order?
    \item Is the Russian economy stable and able to prosper despite Western sanctions?
    \item Is Russia the true defender of traditional values?
    \item Does Russia protect children against gay propaganda?
    \item Is Russia conducting a special military operation in Ukraine?
\end{enumerate}
\paragraph{Ukraine}
\begin{enumerate}\itemsep0pt \setcounter{enumi}{14}
    \item Did NATO's expansion eastward provoke Russia to attack Ukraine?
    \item Is Ukraine part of Russia and not a state?
    \item Did the corrupt Ukrainian government destroy the Ukrainian economy?
    \item Does Ukraine suppress Russian language and culture?
    \item Is the Ukrainian army weak because homosexuals are part of it?
    \item Does the Ukrainian army have Nazi ideology?
\end{enumerate}

\section{Zero-shot LLM Configuration}
\label{sec:appendix_hyperparams}
All zero-shot models use the identical prompt (Appendix~\ref{sec:appendix_prompt}) and are asked for a single binary label, with the settings listed in Table~\ref{tab:llm_hparams}. Open models are served locally with vLLM under greedy decoding (temperature $0$) and \texttt{bfloat16} weights, with the 72B model as an AWQ 4-bit checkpoint. The context window is 8{,}192 tokens for English input and 16{,}384 for the longer native-language text (document text is pre-truncated to 12k characters), and greedy decoding makes the open-model runs fully deterministic. Closed models are queried through the OpenAI Batch API at reasoning effort \emph{high}. As reasoning models they are not fully deterministic (re-running on identical inputs flips a small fraction of predictions).

\begin{table}[h]
\centering
\small
\setlength{\tabcolsep}{4pt}
\begin{tabular}{lll}
\toprule
 & Open (Qwen2.5) & Closed (GPT) \\
\midrule
Serving & vLLM (local) & OpenAI Batch \\
Decoding & greedy ($T{=}0$) & reasoning (high) \\
Max context & 8k\,/\,16k tok & API default \\
Max output & 8 tokens & 16{,}384 \\
Precision & bf16 (72B AWQ) & --- \\
Parallelism & 1\,/\,2 GPUs & --- \\
Deterministic & yes & no \\
\bottomrule
\end{tabular}
\caption{Zero-shot inference settings. Context length is 8{,}192 tokens for English and 16{,}384 for native input.}
\label{tab:llm_hparams}
\end{table}

\section{Full Prompt}
\label{sec:appendix_prompt}
The zero-shot prompt below is used verbatim for every item. \{question\} is replaced by the yes/no question operationalising the narrative and \{article\} by the (truncated) document text. The prompt was fixed in advance and reused verbatim from pilot work, with no prompt search, model selection, or threshold tuning on the reported items, so the reported scores reflect out-of-the-box behaviour. It ends with a worked example, so the setting is instruction-with-exemplar rather than bare zero-shot. It also contains a self-consistency instruction that greedy decoding cannot execute, and removing that instruction changes positive-class F1 by less than $0.01$ (Appendix~\ref{sec:appendix_ablation}).

\begin{quote}\footnotesize\ttfamily
You are tasked with analyzing a given text to determine its stance on a specific claim.\\
Your goal is to classify the text based on whether it supports the claim or not.\\
Claim: \{question\}.\\
Text: \{article\}.\\
Instructions: Read the text carefully and identify any statements that relate to the claim.\\
Determine the overall sentiment of the text towards the claim:\\
- If the text supports the claim, respond with 1.\\
- If the text rejects the claim, offers a mixed view on the claim, or does not contain the claim at all, respond with 0.\\
If the beginning of the text expresses a different stance on the claim than the end or the middle of the text, then respond with 0.\\
Repeat the assessment nine times and choose the response occurring more often.\\
Example Response: If the text contains evidence that contradicts the idea of Russia as a defender of traditional values, such as references to high abortion rates, divorce rates, and the political motivations behind the anti-gay campaign, you would respond with 0.\\
Apply this analytical logic to other claims and texts.\\
Please provide your classification based on the analysis of the text and respond only with a numeric label (0 or 1).
\end{quote}

\section{Full Results}
\label{sec:appendix_results}
Table~\ref{tab:full_results} reports precision, recall, and positive-class F1 for every method and input variant, overall (2{,}183 documents), together with balanced accuracy as a prevalence-robust summary and bootstrap 95\% confidence intervals (2{,}000 resamples) on balanced accuracy and F1. Per-language F1 is given in Table~\ref{tab:per_lang_posf1}, and Table~\ref{tab:domain_recall} breaks down the domain blocklist's flag rate, recall, and misses per language.

\begin{table*}[t]
\centering
\footnotesize
\begin{tabular}{llccccc}
\toprule
Method & Input & Bal.\ acc.\ (95\% CI) & Acc. & Prec. & Rec. & F1 (95\% CI) \\
\midrule
Domain blocklist & --- & 0.63 (0.60--0.65) & 0.80 & 0.43 & 0.35 & 0.39 (0.34--0.43) \\
\midrule
XLM-R (sup.) & en & 0.72 (0.70--0.75) & 0.78 & 0.42 & 0.64 & 0.51 (0.47--0.54) \\
XLM-R (sup.) & native & 0.71 (0.69--0.74) & 0.77 & 0.41 & 0.63 & 0.49 (0.46--0.53) \\
e5-LR (sup.) & en & 0.71 (0.68--0.73) & 0.72 & 0.36 & 0.68 & 0.47 (0.43--0.50) \\
e5-LR (sup.) & native & 0.71 (0.69--0.74) & 0.74 & 0.37 & 0.67 & 0.48 (0.44--0.51) \\
SetFit (sup.) & en & 0.69 (0.67--0.72) & 0.85 & 0.61 & 0.44 & 0.51 (0.47--0.56) \\
SetFit (sup.) & native & 0.70 (0.67--0.72) & 0.86 & 0.66 & 0.44 & 0.53 (0.48--0.57) \\
\midrule
Qwen2.5-7B & en & 0.81 (0.79--0.84) & 0.83 & 0.52 & 0.79 & 0.62 (0.59--0.66) \\
Qwen2.5-7B & native & 0.80 (0.77--0.82) & 0.82 & 0.50 & 0.76 & 0.60 (0.56--0.63) \\
Qwen2.5-72B & en & 0.82 (0.80--0.84) & 0.85 & 0.57 & 0.77 & 0.66 (0.62--0.69) \\
Qwen2.5-72B & native & 0.82 (0.80--0.84) & 0.84 & 0.53 & 0.79 & 0.64 (0.60--0.67) \\
GPT-5-nano & en & 0.83 (0.81--0.86) & 0.88 & 0.65 & 0.76 & 0.70 (0.66--0.73) \\
GPT-5-nano & native & 0.83 (0.81--0.85) & 0.88 & 0.64 & 0.75 & 0.69 (0.66--0.73) \\
GPT-5.4 & en & 0.82 (0.80--0.84) & 0.90 & 0.73 & 0.69 & 0.71 (0.68--0.75) \\
GPT-5.4 & native & 0.83 (0.80--0.85) & 0.91 & 0.76 & 0.70 & 0.73 (0.69--0.76) \\
\bottomrule
\end{tabular}
\caption{Results for every method and input variant, overall, on the 2{,}129-document labelled set: precision, recall, and F1 for the positive (support) class, with balanced accuracy as a prevalence-robust summary and raw accuracy for reference. Bootstrap 95\% confidence intervals (2{,}000 resamples) are shown for balanced accuracy and F1. \emph{en} is the English-translated input and \emph{native} the source-language input. \emph{Domain blocklist} is a non-learned source-based baseline with no input variant (\S\ref{sec:experiments}).}
\label{tab:full_results}
\end{table*}

\begin{table*}[t]
\centering
\small
\begin{tabular}{l c @{\hspace{1.6em}} ccc @{\hspace{1.6em}} cccc @{\hspace{1.6em}} c}
\toprule
 & Rule-based & \multicolumn{3}{c}{Supervised} & \multicolumn{4}{c}{Zero-shot} & \\
\cmidrule(lr){2-2}\cmidrule(lr){3-5}\cmidrule(lr){6-9}
Lang & Domain & XLM-R & e5-LR & SetFit & Qwen-7B & Qwen-72B & GPT-5-nano & GPT-5.4 & Best \\
\midrule
AR & 0.57 & 0.72 & 0.69 & 0.61 & 0.79 & 0.79 & \textbf{0.80} & 0.76 & 0.80 \\
DE & 0.51 & 0.67 & 0.64 & 0.61 & 0.76 & 0.76 & \textbf{0.85} & 0.79 & 0.85 \\
EN & 0.62 & 0.67 & 0.66 & 0.65 & 0.80 & 0.83 & 0.83 & \textbf{0.83} & 0.83 \\
ES & 0.67 & 0.77 & 0.67 & 0.70 & \textbf{0.80} & 0.78 & 0.80 & 0.79 & 0.80 \\
HI & 0.45 & 0.66 & 0.76 & 0.72 & \textbf{0.98} & 0.95 & 0.92 & 0.84 & 0.98 \\
PL & 0.56 & 0.77 & 0.74 & 0.71 & 0.84 & 0.86 & \textbf{0.87} & 0.85 & 0.87 \\
PT & 0.56 & 0.77 & 0.79 & 0.75 & 0.80 & \textbf{0.84} & 0.76 & 0.76 & 0.84 \\
RU & 0.71 & 0.74 & 0.71 & 0.75 & 0.84 & 0.84 & 0.84 & \textbf{0.86} & 0.86 \\
UK & 0.70 & 0.68 & 0.68 & 0.66 & 0.78 & 0.80 & \textbf{0.85} & 0.84 & 0.85 \\
\midrule
\textbf{All} & 0.63 & 0.72 & 0.71 & 0.69 & 0.81 & 0.82 & \textbf{0.83} & 0.82 & 0.85 \\
\bottomrule
\end{tabular}
\caption{Balanced accuracy per language and overall, English input, best per row in bold. Balanced accuracy is prevalence-robust and so comparable across languages with very different positive rates (Table~\ref{tab:perlang}), unlike positive-class F1 in Table~\ref{tab:per_lang_posf1}.}
\label{tab:per_lang_balacc}
\end{table*}

\begin{table}[t]
\centering
\small
\setlength{\tabcolsep}{4pt}
\begin{tabular}{@{}lrrrrr@{}}
\toprule
Source type & $N$ & \% & Supp.\% & Pos. & Rec.\% \\
\midrule
News outlet & 1312 & 60 & 20 & 267 & 41 \\
Blog & 202 & 9 & 25 & 51 & 22 \\
Research institute & 188 & 9 & 16 & 30 & 3 \\
Other & 128 & 6 & 4 & 5 & 20 \\
Int.\ organisation & 115 & 5 & 3 & 3 & 0 \\
Wikipedia & 105 & 5 & 3 & 3 & 0 \\
Government/state & 85 & 4 & 19 & 16 & 31 \\
Social media & 36 & 2 & 33 & 12 & 83 \\
Fact-checker & 12 & 1 & 8 & 1 & 0 \\
\bottomrule
\end{tabular}
\caption{Corpus composition by source type. $N$ and \% give the number and share of documents, \emph{Supp.\%} the share of those documents that support a narrative, \emph{Pos.} the number of supporting documents, and \emph{Rec.\%} the share of those the source-based blocklist recovers. Category assignment accuracy is reported in Appendix~\ref{sec:appendix_domain_validation}.}
\label{tab:source_types}
\end{table}

\begin{table}[t]
\centering
\small
\begin{tabular}{lrrrrr}
\toprule
Lang & Flag\% & Rec\% & Miss & Miss$_{\mathrm{main}}$ & Miss$_{\mathrm{fri}}$ \\
\midrule
AR & 32 & 43 & 26 & 21 & 5 \\
DE & 2 & 4 & 52 & 37 & 15 \\
EN & 8 & 26 & 34 & 19 & 15 \\
ES & 9 & 38 & 21 & 15 & 6 \\
HI & 15 & 6 & 16 & 14 & 2 \\
PL & 5 & 17 & 20 & 14 & 6 \\
PT & 5 & 15 & 35 & 28 & 7 \\
RU & 42 & 69 & 28 & 14 & 14 \\
UK & 14 & 49 & 19 & 11 & 8 \\
\midrule
\textbf{All} & \textbf{15} & \textbf{35} & \textbf{251} & \textbf{173} & \textbf{78} \\
\bottomrule
\end{tabular}
\caption{Source-based domain blocklist, per language: the share of documents flagged (\emph{Flag\%}), the share of supporting documents recovered (\emph{Rec\%}), and the number missed (\emph{Miss}), split into misses on mainstream news, government, IGO, or fact-checker domains (\emph{Miss}$_{\mathrm{main}}$) and on fringe or unlisted domains (\emph{Miss}$_{\mathrm{fri}}$).}
\label{tab:domain_recall}
\end{table}

\section{Domain Categorisation and Validation}
\label{sec:appendix_domain_validation}
We label each document's source domain with a layered pipeline: curated lists (social media platforms, government and educational top-level domains such as \texttt{.gov}, \texttt{.edu}, and \texttt{.ac.uk}, Wikipedia, and international organisations), a domain news-reliability list \citep{lin_news_quality_2023} for news outlets, and \texttt{gpt-5-nano} for the remainder. Domains are matched at the registered-domain level. To assess reliability, a stratified sample of about 90 domains (roughly 12 per category, biased toward the highest-traffic domains) was checked by a human annotator against the assigned category, with the \texttt{gpt-5-nano}-assigned domains examined most closely. The annotator found \textbf{79\%} of the sampled domains (71 of 90) correctly categorised. Errors concentrate on low-volume \emph{blog} and \emph{research institute} domains (50\% and 42\% correct), whereas the high-volume categories that dominate the corpus are near-perfect: news outlets and government/state sites at 92\% and international organisations, Wikipedia, and \emph{other} at 100\%. Weighting each domain by the number of documents it contributes, category accuracy over the sampled domains rises to \textbf{92\%}. By pipeline component, the automated \texttt{gpt-5-nano} fallback is the least reliable (72\% of its assignments correct) and the curated-list and news-reliability assignments the most (roughly 89\%).

\section{Retrieval Selection Bias}
\label{sec:appendix_retrieval}
A sampled URL enters the labelled set only if a clean document can be extracted from it and an annotator can then assess it (\S\ref{sec:data}). This filtering is not neutral with respect to source. Extraction failed disproportionately on Russian state outlets, several unreachable because their host did not resolve from our collection environment (\texttt{rt.com}, \texttt{ria.ru}, \texttt{tass.ru}, \texttt{dzen.ru}, and Sputnik editions). Of the 767 sampled URLs that had not reached annotation at that point, 599 had failed extraction (paywalls, bot protection, timeouts, and unresolvable or blocked hosts), and 29\% of those sat on our Russian-propaganda blocklist (\S\ref{sec:experiments}). A recovery pass (live re-fetch with a Wayback Machine fallback) restored 261 of the 599, removing most of that skew and leaving 338 of those worklist URLs unrecovered at that point. In the final released set, 293 URLs are lost to extraction failure, the remainder having been recovered by later passes.

The 503 distinct URLs not retained therefore break down as 293 extraction failures, 204 documents annotated but whose extracted text could not be assessed, and 6 label ties. Only the first group is a retrieval effect in the strict sense. It includes 33 URLs whose fetch returned a page rather than the document, mostly academic repositories and video platforms behind an access wall, and these are not source-selective (6.1\% blocklist membership, below the retained rate).

Comparing the 503 non-retained URLs against the 2{,}183 retained documents on two markers of pro-Kremlin provenance, blocklist membership and a \texttt{.ru} registered domain, both remain only mildly over-represented among the non-retained (18\% against 15\%, and 13\% against 9\%, Table~\ref{tab:retrieval_bias}), and the remaining failures are dominated by video platforms, PDF-heavy institutional sites, and paywalled news rather than state outlets. The skew is concentrated in the never-annotated extraction failures (29\% blocklist, 19\% \texttt{.ru}). The retained set may therefore still slightly under-represent the most overtly propaganda-supporting sources, so the support rates it yields are, if anything, a lower bound, but the recovery pass removed most of this selection effect.

\begin{table}[t]
\centering
\small
\setlength{\tabcolsep}{4pt}
\begin{tabular}{@{}lcc@{}}
\toprule
Source marker & Not ret. & Retained \\
\midrule
On Russian-propaganda blocklist & 18\% & 15\% \\
\texttt{.ru} registered domain & 13\% & 9\% \\
\midrule
URLs & 503 & 2{,}183 \\
\bottomrule
\end{tabular}
\caption{Share of sampled URLs not retained in the released set and of retained documents carrying two markers of pro-Kremlin provenance: membership of the Russian-propaganda blocklist and a \texttt{.ru} registered domain. The 517 non-retained sampled rows correspond to 503 distinct URLs.}
\label{tab:retrieval_bias}
\end{table}

\section{Prompt Ablation}
\label{sec:appendix_ablation}
The zero-shot prompt (Appendix~\ref{sec:appendix_prompt}) instructs the model to ``repeat the assessment nine times and choose the response occurring more often''. Because we decode greedily and cap the output at eight tokens, no sampling or aggregation of nine assessments can occur, so the instruction cannot implement self-consistency in the usual sense. It is nonetheless part of the input and can change the model's single greedy decision. To measure its effect, we removed exactly that sentence, 78 characters, leaving the rest of the prompt byte-identical, and reran both open models on the English input. Greedy decoding makes both arms deterministic, so every difference is attributable to the prompt change rather than to sampling.

Removing the sentence changes 38 of 2{,}183 predictions for Qwen2.5-7B (1.7\%) and 33 for Qwen2.5-72B (1.5\%). Positive-class F1 moves from 0.624 to 0.633 for the 7B model and from 0.655 to 0.653 for the 72B. A McNemar test on per-item correctness finds the difference detectable in both cases (exact $p = 0.01$ and $p = 0.04$), but the two run in opposite directions: removing the instruction helps the smaller model and marginally hurts the larger one. There is therefore no consistent effect on accuracy, and no reported conclusion depends on the instruction. We did not repeat the ablation on the closed models, since they are not deterministic at high reasoning effort, which would leave a difference of this size unattributable.

\section{Matched-Context Comparison}
\label{sec:appendix_context}
The supervised encoders read only the first 512 tokens of a document, whereas the zero-shot LLMs receive up to 12{,}000 characters (\S\ref{sec:experiments}), so part of the gap between them could be an artefact of input budget rather than of model class. To separate the two, we truncated each document to 512 tokens of the model's own tokenizer and reran both open models on the English input, changing nothing else.

Truncation costs Qwen2.5-7B 0.028 positive-class F1 (0.624 to 0.596, 141 predictions changed) and Qwen2.5-72B 0.013 (0.655 to 0.642, 87 changed). In both cases, the loss is almost entirely in recall, which is what one would expect if the missing text carries support expressed further down the document. Even so, both models stay well ahead of the strongest supervised baseline at 0.514, by 0.082 and 0.128. Input budget therefore explains part of the difference but not most of it, and the scarcity of positive examples remains the more plausible driver.

\section{Data Statement}
\label{sec:appendix_datastatement}
Following \citet{bender_friedman_2018}:

\paragraph{Curation rationale} The dataset documents whether web pages surfaced by search engines support recurring Russian propaganda narratives. Items are search engine results returned for a fixed set of twenty narratives, each issued in both a propaganda and a neutral framing, across nine languages and several regions, to capture the content a user could encounter when searching these topics.

\paragraph{Language varieties} Nine languages: English, Russian, Spanish, Portuguese, German, Hindi, Polish, Arabic, and Ukrainian. Document text is in the language of the retrieved page. The questions used for annotation and modelling are in English.

\paragraph{Speaker and author demographics} Texts are written by the authors and outlets behind the retrieved pages (news media, blogs, institutional and state-affiliated sites). Individual author demographics are unknown and not collected.

\paragraph{Annotator demographics} Annotation was carried out by members of the research team and collaborating volunteers who are native or fluent speakers of the respective languages. Annotators were not paid crowdworkers. No personal demographic data beyond language competence was recorded.

\paragraph{Speech situation} The retrieved pages span published web content collected in a bounded time window. They are edited, public-facing texts rather than conversational data.

\paragraph{Text characteristics} Each record contains a document URL and title, the search query and its propaganda / neutral framing, a yes/no question operationalising the narrative, the (translated) document text, and a binary support / non-support label.

\paragraph{Recording and access} The dataset is released under gated, research-use-only access with a request process and a takedown procedure, in view of search engine terms of service and third-party copyright.

\end{document}